# Learning and Simulating Building Evacuation Patterns for Enhanced Safety Design Using Generative Models

Jin Han[1], Zhe Zheng[1,2], Yi Gu[1], Jia-Rui Lin[1,3,*], Xin-Zheng Lu[1]

1. Department of Civil Engineering, Tsinghua University, Beijing, 100084, China

2. Dept. of Investment and Technology Innovation, Wuliangye Yibin Co., Ltd.

3. Key Laboratory of Digital Construction and Digital Twin, Ministry of Housing and Urban-Rural Development, Beijing, 100084, China

*Corresponding author, E-mail: lin611@tsinghua.edu.cn

**Abtract:**

Evacuation simulation is essential for building safety design, ensuring properly planned evacuation routes. Thus, this study proposes DiffEvac, an innovative method for efficient evacuation simulation and building safety design. Unlike traditional simulations that rely on extensive parameter modeling, DiffEvac uses Generative Models (GMs) to learn evacuation patterns, enabling faster iteration in early design stages. Initially, a dataset of 399 diverse functional layouts and corresponding evacuation heatmaps of buildings was established. Then, decoupled feature representation is proposed to embed physical features like layouts and occupant density for GMs. Finally, a diffusion model based on image prompts is proposed to learn evacuation patterns from simulated evacuation heatmaps. Compared to existing research, DiffEvac achieves up to a 37.6% improvement in SSIM, 142% in PSNR, and delivers results 16 times faster—cutting simulation time to 2 minutes. Case studies demonstrate that DiffEvac enhances design iterations and provides innovative pathways for intelligent building safety optimization.

**Keywords:**

Efficient evacuation simulation; Generative neural networks; Diffusion model; decoupled feature representation; Surrogate model

# 1 Introduction

In complex large-scale buildings, pedestrian traffic is often extremely dense, and the internal structure is frequently intricate, posing significant challenges to safe evacuation. Research shows that there were at least 41 global crowd stampedes, resulting in 2,683 deaths and 3,014 injuries between 2007 and 2019 (Liu et al., 2019). Furthermore, in major stampede incidents both domestically and internationally, approximately 70% of fatalities were attributed to inadequate evacuation and crowd congestion (Li et al., 2022). Therefore, effective evacuation in building design is essential for safeguarding occupants during emergencies such as fires and earthquakes (Li et al., 2024).

Effective evacuation simulation not only enhances response efficiency during emergencies but also provides a scientific foundation for building design and safety assessments. As a predictive and analytical tool, evacuation simulation facilitates the understanding of complex evacuation behaviors, enhances building design, and refines emergency plans. Consequently, it has garnered significant attention from both academic researchers and industry practitioners. Existing evacuation simulations can be categorized into three types: macro-level models, micro-level models, and hybrid simulations, depending on whether they focus on overall crowd behavior or individual interactions. Macro-level simulation approaches treat the crowd as a continuous fluid, using fluid dynamics equations to describe crowd movement. These methods are suitable for evacuation simulations of large-scale crowds due to their high computational efficiency, but they struggle to reflect individual behavioral differences. Common macroscopic models include the social force model and the lattice gas model, among others (Klote & Hadjisophocleous, 2008; Helbing et al., 2000; Helbing et al., 2003). With advancements in computational power and data accessibility, researchers have increasingly focused on micro-level simulations, proposing more complex and refined models, including agent-based models and cellular automata models (Cotfas et al., 2022; Alac et al., 2023; Lim et al., 2023; Fu et al., 2015; Gao et al., 2022). These models, by treating each evacuee as an autonomous agent, can simulate individual behavioral differences and interactions among them, such as panic, herding, and competition (Senanayake et al., 2024). Hybrid simulation methods combine the advantages of macro-level and micro-level approaches, enhancing simulation accuracy while maintaining computational efficiency. For instance, macroscopic simulations can be applied to large-scale areas, while microscopic simulations are used in critical regions (Xiong et al., 2013; Serena et al., 2023). Although these models can provide a precise simulation of evacuation patterns, they also demand more refined modeling and a larger number of input parameters, which are time-consuming and labor-intensive.

However, design time is limited, and architects often face numerous design alternatives during the review process (Ritter et al., 2015). In addition, sufficient parameters for detailed simulation are usually not available in the early design stage, but early consideration of the building layout's impact on evacuation can reduce revisions and rework (Pelechano et al., 2008; Østergård et al., 2016). Obviously, traditional evacuation simulation methods, which typically rely on detailed, computationally intensive simulations, are inadequate for the review process that demands rapid iteration and adjustment. These methods also pose challenges in schematic design stages, which are based solely on basic architectural sketches that have yet to be developed or detailed. Therefore, finding ways to simplify the application of evacuation simulation methods through computer technology to enhance safety design is an urgent problem that needs to be addressed.

With the rapid development of artificial intelligence (AI) technologies capable of learning from existing data and knowledge, deep neural networks have demonstrated powerful abilities in nonlinear and fuzzy learning (Salehi & Burgueño, 2018), and image generation (Chen et al., 2024). These methods are expected to learn the potential mapping relationships from architectural sketches to evacuation heatmaps, thereby achieving efficient evacuation simulation. Although some research, such as Nourkojouri et al. (2023), has begun exploring the use of image generation algorithms for rapid evacuation simulation and evaluation, their study primarily employed the Conditional GAN (Generative Adversarial Networks) model for preliminary experiments, which is weaker in generalization and image generation detail compared to newer algorithms like diffusion models (Gu et al., 2024). Besides, their research also showed poor performance on irregular floor plans, highlighting the need to analyze various image generation models, as they may offer advantages in quality and stability, better supporting rapid evaluations of different building layouts.

Therefore, this study proposes DiffEvac, a diffusion model based on the image prompt method with decoupled feature representation for learning building evacuation patterns to quickly generate evacuation flow accumulation heatmaps (hereinafter referred to as evacuation heatmaps). The surrogate model is designed to replace some functions of complex models with a simpler, more computationally efficient alternative. Since this study uses a cost-effective deep learning model to substitute for a more expensive evacuation simulation software, the developed model will be referred to as the evacuation surrogate model in the following sections. The remainder of this paper is organized as follows. Section 2 reviews related work and outlines existing research gaps. Section 3 details the construction of the dataset and the evacuation surrogate model. Section 4 compares DiffEvac with other commonly used methods. Section 5 demonstrates the practical application value of DiffEvac through case studies. Finally, Section 6 concludes this research.

# 2 Related work

## 2.1 Modeling-based evacuation simulation method

Based on in-depth research into phenomena and patterns related to disasters such as fires and earthquakes, it is possible to achieve a quantitative description and analysis of human behavior, environmental characteristics, and disaster impacts. This, in turn, allows for effective prediction and optimization of building evacuation performance, providing a scientific basis for architectural design and emergency management. Existing evacuation simulations can be categorized into three types: macro-level models, micro-level models, and hybrid simulations, depending on whether they focus on overall crowd behavior or individual interactions. Macro-level simulation approaches treat the crowd as a continuous fluid, using fluid dynamics equations to describe crowd movement. Common macroscopic models include the social force model and the lattice gas model, among others. Klote and Hadjisophocleous (2008) modeled overall crowd movement as a fluid-like motion to estimate evacuation times. Helbing et al. (2000) integrated social force models by considering avoidance behaviors and microscopic actions to predict evacuation time. Helbing et al. (2003) introduced the Lattice Gas Model to discretize space into a grid, simulating individual movement through probabilistic rules. These methods are suitable for evacuation simulations of large-scale crowds due to their high computational efficiency, but they struggle to reflect individual behavioral differences.

With advancements in computational power and data accessibility, researchers have increasingly focused on micro-level simulations, proposing more complex and refined models, where individual agents serve as the basic unit and are endowed with autonomous decision-making capabilities. Agent-based models (ABMs) have gained attention due to their ability to account for individual differences and uncertainties in decision-making processes (Lim et al., 2023). Cotfas et al. (2022) and Alac et al. (2023) have introduced agents to simulate individual behavior characteristics during evacuation, such as panic responses, path selection, and avoidance, thereby enabling reliable optimization of egress locations and crowd evacuation paths. Additionally, Fu et al. (2015) and Gao et al. (2022) proposed a velocity-adjusted cellular automata model that more accurately simulates changes in the number of remaining people over time by considering individual walking speed variations.

In contrast, hybrid simulation methods use macroscopic models at the global level to enhance efficiency, while employing microscopic models in critical areas (such as exits and staircases) to capture finer details (Xiong et al., 2013; Serena et al., 2023). This approach performs well in handling complex scenarios, but model construction and parameter adjustment can be relatively complex. The choice of an appropriate simulation method requires a trade-off based on the specific scenario, research objectives, and computational resources.

Based on these developments, a range of disaster evacuation simulation software has emerged, including Evacnet, AnyLogic, Wayout, Steps, Pedgo, Simulex, Pathfinder, Building Exodus, and MassMotion (Senanayake et al., 2024). Utilizing these tools, numerous studies have been conducted on evacuation under fire scenarios. Specific examples include evaluations of evacuation effectiveness using BIM and AnyLogic (Sun & Turkan, 2019, 2020) and assessments of building evacuation performance based on BIM and EvacuatioNZ (Dimyadi et al., 2017). Although these models and software can provide a precise simulation of evacuation patterns, they require more detailed specifications and a greater number of input parameters, making them time-consuming and labor-intensive.

However, during the review and adjustment process, architects often face numerous design alternatives. Additionally, in the schematic design stage, it is crucial to preliminarily determine factors such as the layout and location of evacuation routes for safe design. Considering the impact of building layout on evacuation early in the design process, and thus avoiding layouts that hinder evacuation, can significantly reduce the frequency of revisions and rework. Existing evacuation simulation methods typically rely on detailed building parameters and three-dimensional simulation models, making the complex modeling process and lengthy simulation times (Nourkojouri et al., 2023), which is unsuitable for the schematic design stage with only architectural sketches and the need for evaluating numerous design options. Therefore, how to leverage computer technology to simplify the application of evacuation simulation methods is an urgent issue that needs to be resolved.

**2.2 Machine learning-based evacuation simulation method**

Research shows that when trained on historical evacuation data, machine learning has the potential to learn underlying evacuation patterns, such as Support Vector Machines (SVM) and Random Forests (Wang et al., 2019; Zhao et al., 2020). Based on this, Zhu et al. (2023) utilized machine learning and discrete choice models to discuss evacuation performance under various factors, such as building attributes and residents' familiarity with the building. Beyond behavior prediction, some studies have used machine learning algorithms to optimize evacuation strategies, aiming to minimize evacuation time and risk. For example, von Schantz & Ehtamo (2022) combined numerical simulations with genetic algorithms to evaluate and enhance the safety and efficiency of different evacuation plans. However, while these studies demonstrate the potential of machine learning in evacuation simulation, they also face the challenge of requiring detailed design and population parameters.

In recent years, with the rapid growth in computational power and data volume, Deep Neural Networks (DNNs) have demonstrated exceptional performance in handling complex nonlinear relationships and large-scale data (Salehi & Burgueño, 2018). By processing extensive datasets, DNNs are capable of analyzing and processing image and video data in real time, making them suitable for monitoring and analyzing behavioral patterns during evacuations. For instance, Haque et al. (2020) utilized Convolutional Neural Networks (CNNs) to assess crowd density appropriations in different regions from surveillance videos, thereby offering immediate evacuation guidance and preventing stampede incidents.

Generative AI is one of the important technologies in the field of deep learning, demonstrating impressive capabilities in both fuzzy learning and image generation (Chen et al., 2024). Notable examples include Generative Adversarial Networks (GANs), U-Net models, and Diffusion Models. GANs, introduced by Ian Goodfellow et al. (2014) in 2014, represent a milestone in image generation. They innovatively frame the generative problem as a game between a generative network and a discriminative network, showcasing strong end-to-end generation capabilities. Based on this approach, Liao et al. (2021), Fei et al. (2022), Fu et al. (2023), and Han et al. (2024) utilized GANs to learn the underlying design patterns in structural drawings, facilitating the intelligent design of shear walls and steel frame-braced structures. Additionally, the U-Net model, proposed by Ronneberger et al. (2015), is a deep generative neural network featuring an encoder-decoder architecture with skip connections. It is widely used for image segmentation and generation tasks due to its ability to achieve high accuracy with less training data. Jiang et al. (2022) utilized U-Net for façade orthoimage pixelwise segmentation. In addition, Diffusion Models, a more recent type of generative model, produce data by progressively refining random noise into the desired output, demonstrating exceptional performance in image generation in recent years. Gu et al. (2024) applied Diffusion Models to architectural structural design and demonstrated superior performance compared to GANs in capturing engineering design features and optimizing performance metrics. These studies collectively confirm that generative neural networks can learn architectural spatial layouts and topological features by training on existing drawings and knowledge. This suggests the potential for these models to learn the underlying mapping from architectural sketches to evacuation heatmaps, thus enabling efficient evacuation simulation.

However, research in this area remains limited. Although some research, such as Nourkojouri et al. (2023), has begun exploring the use of image generation algorithms for rapid evacuation simulation and assessment. Their study, however, primarily employed the Conditional GAN model for initial experimentation, which is weaker in generalization and image generation detail compared to newer algorithms like diffusion models (Gu et al., 2024). Besides, their research also showed poor performance on irregular floor plans, highlighting the need to analyze various image generation models.

## 2.3 Research gaps

Although previous research has made significant contributions to evacuation simulation, there are two knowledge gaps that this paper aims to address: (1) while many studies can provide a precise simulation of evacuation patterns, they require more refined modeling and a larger number of input parameters, making them time-consuming and labor-intensive. Consequently, they are unsuitable for review processes that demand rapid design iterations and the schematic design stage, which relies solely on architectural sketches that have not yet been developed and detailed. (2) existing research on evacuation simulation using generative models (GMs) showed poor performance on irregular floor plans, highlighting the need to analyze various image generation models.

In light of the limitations of traditional methods in addressing rapid design and review needs, this study proposes a novel surrogate model, DiffEvac, that uses GMs to learn building evacuation patterns. By facilitating rapid simulations in place of time-consuming conventional approaches, DiffEvac helps predict critical evacuation points during schematic design and review stages. This capacity for swift iteration and layout refinement underscores the pressing need for more efficient safety design solutions.

## 3 Methodology

To address the issues of time-consuming and high demands for detailed architectural information in existing evacuation simulation methods, this paper proposes DiffEvac, an efficient evacuation simulation method based on deep-learning image generation algorithms, as illustrated in Figure 1. We aim for DiffEvac to learn evacuation patterns from the datasets generated by Pathfinder software, enabling it to perform rapid simulations as an alternative to the time-consuming simulations typically required by Pathfinder.

Since constructing and training the surrogate models requires relevant datasets, we first develop an evacuation dataset for training deep-learning models, comprising an input image set of functional layout drawings and corresponding evacuation heatmaps as a ground-truth set. This process unfolds in three stages: preprocessing of building floor plans, Pathfinder-based evacuation simulation, and alignment of the evacuation dataset. Detailed procedures for dataset construction are described in Section 3.1. Subsequently, we proposed a diffusion model based on the image prompt method and a decoupled feature representation approach to enhance the model's simulation performance. Specifically, since the widely used RGB three-channel representation cannot directly reflect physical features like room layout and occupant density, this study proposes a decoupled feature representation method. Additionally, we proposed a diffusion model based on the image prompt method to learn evacuation patterns from simulated evacuation heatmaps generated by Pathfinder software, as illustrated in Section 3.2. Finally, to validate the superiority of the proposed DiffEvac in evacuation simulation, we compare it with other commonly used feature representations and state-of-the-art GMs (Section 3.3), and apply it to case studies to demonstrate demonstrates its practical value (Section 5).

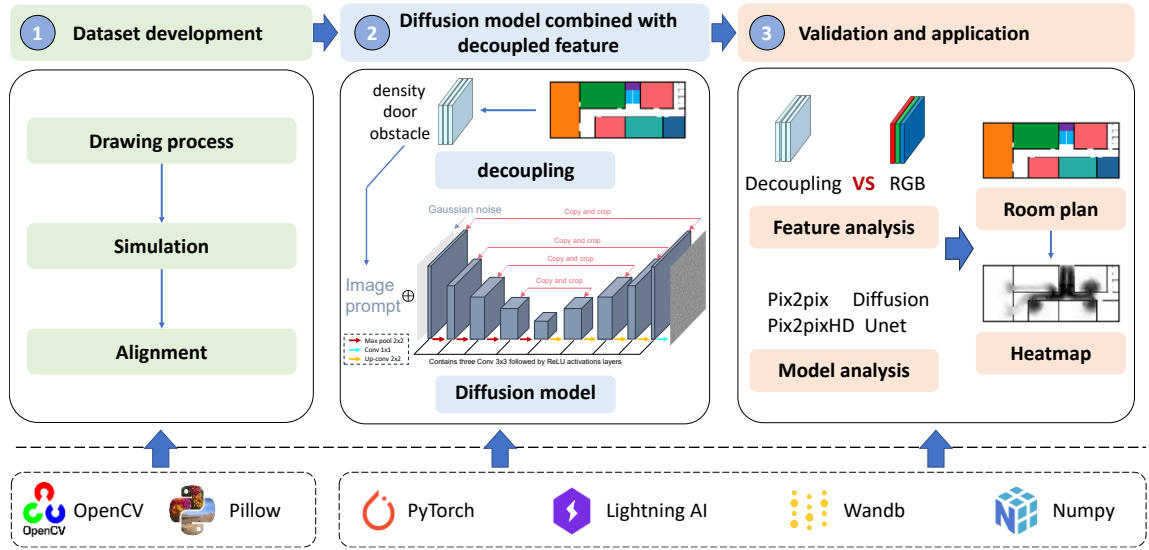

Figure 1 Methodology for rapid simulation of evacuation based on image generation algorithms

## 3.1 Drawing processing and dataset construction

This section aims to establish the evacuation dataset for training deep-learning models and will provide a detailed description of the three sub-steps: processing of building floor plans, Pathfinder-based evacuation simulation, and alignment of the evacuation dataset. Figure 2 illustrates the images before and after processing for each sub-step. Firstly, 81 actual office building floor plans with various irregular layouts were collected to enhance the model's generalization. These plans were then cleaned, annotated, and augmentation according to relevant regulations to generate the room's functional layout drawings, which serve as the input images. After that, Pathfinder was used to construct refined models from floor plans and generate evacuation heatmaps, which serve as the ground truth for guiding the model's learning process. Finally, the input images and ground-truth are aligned to assemble the training dataset.

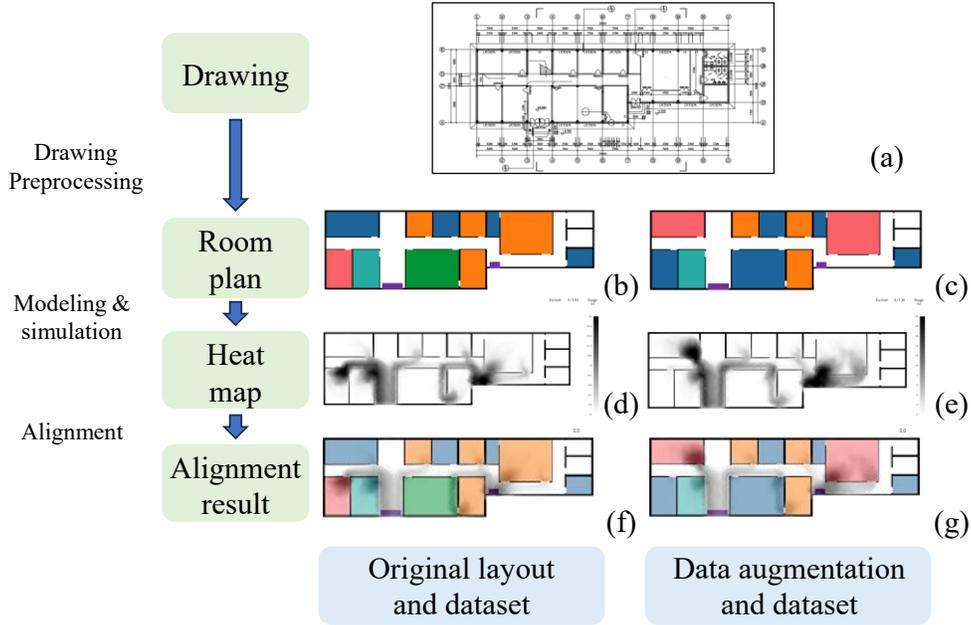

Figure 2 Workflow of drawing processing and dataset construction

### 3.1.1 Drawing preprocessing

Firstly, representative building floor plans are collected. The distribution of occupants varies across different building types: for instance, primary and secondary schools predominantly house minors, nursing homes are mainly occupied by the elderly, and hospitals serve individuals with limited mobility (such as patients and disabled persons). Accurately determining evacuation parameters, such as evacuation speed, for these diverse age groups and health conditions poses challenges and can impact the accuracy of evacuation simulations. In contrast, office buildings typically accommodate adults, with more consistent evacuation parameters. Hence, this study uses office buildings as a case study to develop and validate the proposed evacuation surrogate model. Additionally, Nourkojouri et al. (2023) demonstrated that if a model is trained solely on regular layouts, it will struggle to adapt to diverse building configurations, limiting its generalization ability. Therefore, we collected 81 actual office building floor plans featuring a range of layouts, including rectangular, T-shaped, L-shaped, U-shaped, and other irregular shapes. The quantities for each layout type are shown by the pink bars in Figure 3. These floor plans also vary in the number of rooms and evacuation routes, ensuring a degree of diversity.

Subsequently, the collected building floor plans were cleaned. Specifically, we first removed any drawings that did not pertain to room layouts, such as structural and construction drawings. Then, we eliminated various annotations, text, and redundant lines from the building floor plans, retaining only the walls, windows, and door openings (without indicating door swing directions), and converted them into RGB images. Each drawing was adjusted to fill the image as much as possible, resulting in variable scales. Chang et al. (2021) have shown that crowd density, area capacity, the location where the fires occurred, and exit location are key parameters in evacuation simulations. Since this study focused on evaluating building evacuation performance rather than specific fire scenarios, we assumed all occupants evacuated via the shortest path from their rooms, and the exact location of the fire was not considered. Additionally, while area capacity and exit locations in building floor plans can be directly represented by pixel positions, crowd density cannot be directly visualized, requiring manual annotation. Based on the "Standard for Design of Office Building (JGJ/T67-2019)" (The Ministry of Housing and Urban-Rural Development of the People's Republic of China, 2019) and the "Code for Design of Library Buildings (JGJ 38-2015)" (The Ministry of Housing and Urban-Rural Development of the People's Republic of China, 2015), occupant densities for rooms of different functions were determined, as detailed in Table 1. We used different colors to distinguish room occupancy density and function, embedding crowd density information into the floor plans to ensure the model receives all key parameters. Following these, the building floor plans were cleaned, and annotation was performed according to the color schemes defined in Table 1, resulting in the room's functional layout drawings. For special room types or non-standard building layouts not explicitly covered in the standards, the density can be estimated by the architect, who then selects the closest corresponding color.

Table 1 Occupant densities for rooms of different functions

| Function of room | Density (m²/person) | Color |
| --- | --- | --- |
| Ordinary office | 6 |  |
| Meeting room (with table) | 2 |  |
| Meeting room (no table) | 1 |  |
| Exhibition Hall | 1.43 |  |
| Other region | 9 |  |
| Corridor & Restroom | 0 |  |
| Exit (stairs) | 0 |  |
| Exit (door) | 0 |  |

Finally, data augmentation was performed on the cleaned dataset. Research suggests that deep learning-based image generation models, such as GANs, require hundreds of training samples to achieve optimal performance (Nourkojouri et al., 2023; Liao et al., 2021). However, directly modifying room shapes or rearranging the layout in floor plans for data augmentation is challenging. Given that occupancy density in different rooms is a key parameter and variable in evacuation simulations, we achieved data augmentation by altering room functions and adjusting occupancy densities. Specifically, during the data annotation process, the functions of two or more original rooms were randomly altered and re-annotated according to the defined color schemes, resulting in the creation of four or five distinct room functional layout drawings per original image, as illustrated in Figure 2(b) and Figure 2(c). In total, 399 room functional layout drawings were generated. An overview of the annotated room functional layout drawings dataset and the distribution of quantities by different shapes after data augmentation is shown in Figure 3, illustrating the diverse range of layouts. This variety aids in training models with robust generalization capabilities.

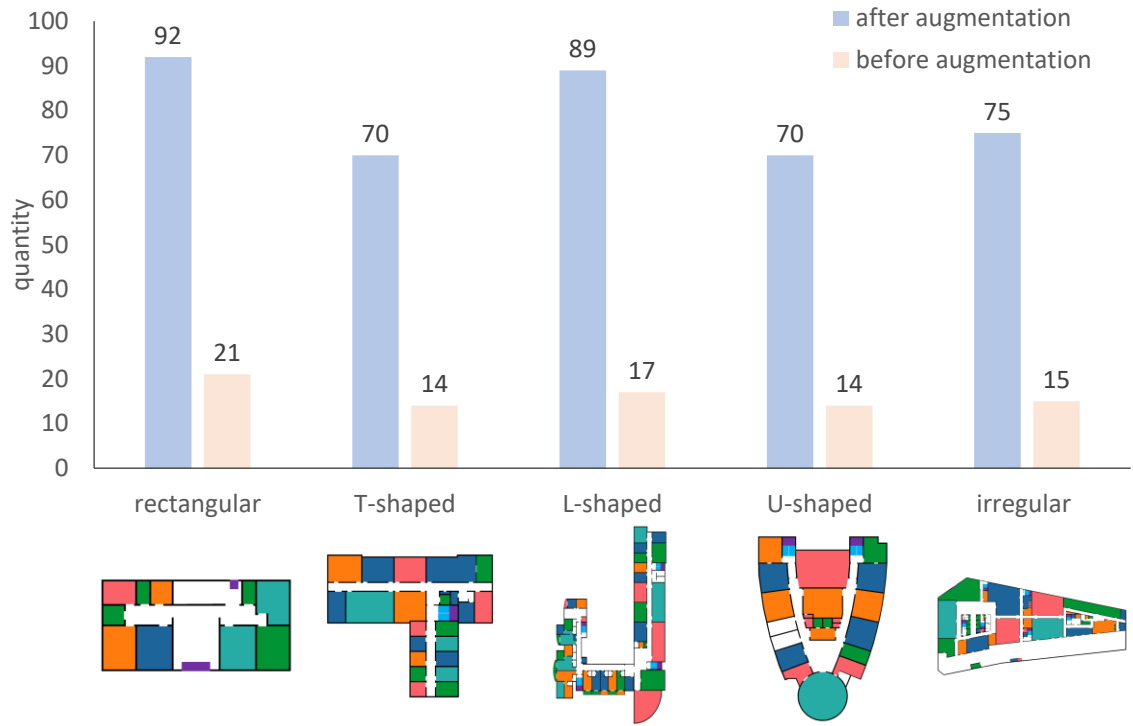

Figure 3 Overview of annotated room functional layout drawings

**3.1.2 Evacuation modeling and simulation**

After obtaining the augmented and annotated drawings, refined models were developed and simulations were completed to generate the ground-truth for guiding the model's learning process. During the simulation, this study set the maximum speed for adults at 1.19 m/s based on the SFPE Handbook of Fire Protection Engineering (Hurley et al., 2016) and existing research (Nourkojouri et al., 2023). Other influencing factors, such as the distribution of people in each room, the types of individuals, and the configuration of emergency exits, were set to the default values provided by the Pathfinder software (Thunderhead Support, 2020). Subsequently, Pathfinder was used to model the room's functional layout drawings and conduct an evacuation simulation analysis. The results of the simulation were post-processed using Pathfinder's post-processing functions to generate evacuation flow accumulation heatmaps (evacuation heatmaps). It should be noted that the evacuation heatmap represents the cumulative time each occupant spends at each point during the evacuation, with darker colors indicating longer cumulative time at that point. These heatmaps can analyze congestion patterns along evacuation routes within the building layout. Specifically, we normalized the simulation results for each drawing to ensure that locations with a gray value of 255 in the evacuation heatmap corresponded to the highest personnel density, while gray values of 0 corresponded to the lowest personnel density. Unlike the RGB-based evacuation heatmaps used in existing studies (Nourkojouri et al., 2023), this study utilized black-and-white single-channel heatmaps to reduce the learning complexity for deep learning models and enhance prediction performance. The evacuation heatmaps represent the cumulative time each occupant spends at each point during the evacuation, with darker colors indicating longer cumulative times at those points. In total, 399 models were created, each with its corresponding evacuation heatmap. Typical evacuation heatmaps are shown in Figure 2(d) and Figure 2(e).

**3.1.3 Alignment of datasets**

Since the room functional layout drawings obtained in Section 3.1.1 and the evacuation heatmaps obtained in Section 3.1.2 were exported from different software, there were pixel-level discrepancies and spatial misalignment between them. Deep learning-based image generation models struggle to learn from misaligned images. To address this, an image alignment algorithm was developed using OpenCV and Pillow. First, the contours of both the room functional layout drawings and the evacuation heatmaps are detected using OpenCV. Next, morphological operations such as dilation and erosion are applied to refine the contours for accurate extraction. Subsequently, these contours are aligned to ensure spatial consistency between the images using Pillow. The results of this alignment process are shown in Figure 2(f) and Figure 2(g). Finally, the images were resized to 256 × 256 pixels to form the final evacuation dataset for model training.

**3.2 Decoupled feature representations**

Existing research commonly maps CAD drawings to the RGB color space by encoding various components with different RGB values. However, this approach can not directly reflect physical features like room layout and occupant density. Consequently, this study proposed a decoupled feature representation method, which separated and encoded different physical information of the drawings into distinct layers. These layers will serve as inputs for the model. Based on the analysis in Section 3.1, crowd density, area capacity, and exit location are key parameters in this study's evacuation simulations. Therefore, the physical information expressed by the room function layout drawing mainly contains three layers: the location of obstacles such as building walls to reflect area capacity, the occupant density within rooms, and the location of exits. These layers can be isolated into separate input channels, resulting in three channels for this study, allowing the model to process and analyze these features more effectively and independently. The representation of these decoupled features is shown in Figure 4.

For the obstacle channel, positions with obstacles are assigned a value of 1, while all other positions are assigned a value of 0. This binary representation allows for the assessment of area capacity by delineating usable and non-usable spaces within the layout. For the room population density channel, the matrix values are filled according to the room locations and types, representing the occupant density, as shown in Table 1. For the exit channel, a value of 1 is assigned to positions where exits are located, and 0 elsewhere. In each layer, the length of consecutive different values represents the size of the corresponding feature, maintaining the same scale as the RGB image, as illustrated in Section 3.1. Compared to RGB representation, decoupled feature representation separates different physical information into distinct channels, avoiding the complexity of mixing all information in RGB. This focused representation allows the model to process simpler and more relevant information, which streamlines computations and enhances efficiency.

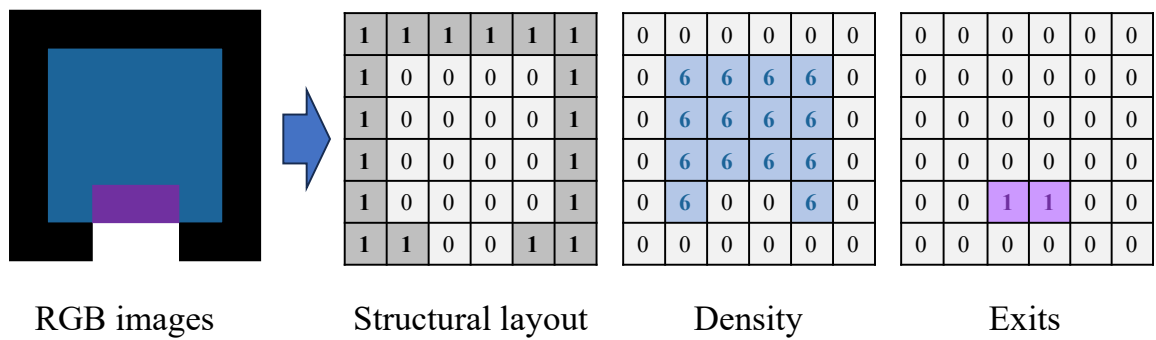

Figure 4 Example of decoupled feature representation

## 3.3 DiffEvac: Improved diffusion model via image prompt

Diffusion models have gained prominence in image generation techniques since 2020 (Dhariwal & Nichol, 2021), training deep neural networks to predict noise and progressively eliminate it to achieve specific generation tasks (Ho et al., 2020). Unlike GANs, diffusion models can provide more detailed and richer results, which is attributed to their iterative noise reduction process (Savinov et al., 2021). For this study, we propose a diffusion model based on the image prompt method, incorporating a denoising approach with an attention mechanism and temporal encoding within a U-Net framework. By learning evacuation heatmaps generated by Pathfinder software in the dataset, the model grasps the patterns of evacuation simulations, enabling it to replace Pathfinder and quickly generate simulation results.

The forward process in a diffusion model is a Markov chain that starts from a real image $x_0$ and gradually adds noise to generate a sequence of noisy images $x_1, x_2, \ldots, x_T$, eventually reaching a fully noisy image $x_T$. This process is defined as:

$$q(x_t|x_{t-1}) = N(x_t; \sqrt{1-\beta_t}x_{t-1}, \beta_t \mathbf{I}) \quad (1)$$

where $\beta_t$ is a noise parameter for time step $t$. In this study, a linear noise schedule is employed, meaning that $\beta_t$ increases linearly from 0.000001 to 0.01 (Dhariwal & Nichol, 2021).

To enable the model to denoise based on physical information, we employed the image prompt method. Specifically, the reverse process generates the image $x_0$ from pure noise $x_T$ combined with either the RGB image or the decoupled feature representation, which is the objective of model training. This process also follows a Markov chain, where the model learns the transition probabilities from $x_t$ to $x_{t-1}$:

$$p_\theta(x_{t-1}|x_t) = N\left(x_{t-1}; \mu_\theta(x_t, t), \sum\nolimits_\theta (x_t, t)\right) \quad (2)$$

where $\mu_\theta(x_t, t)$ and $\Sigma_\theta(x_t, t)$ are the parameters learned by a neural network.

To train the model, the variational lower bound (VLB) is used as the loss function. This aims to minimize the KL divergence between the forward and reverse processes:

$$L_{\text{diffusion}} = \mathbb{E}_q \left[ \sum_{t=1}^{T} D_{KL}\left(q(x_{t-1}|x_t, x_0) \| p_\theta(x_{t-1}|x_t)\right) \right] \quad (3)$$

In this study, a U-Net network with temporal encoding (Dhariwal & Nichol, 2021) was utilized as the denoising model, as depicted in Figure 5. To further explore performance enhancement strategies, the impact of incorporating the attention mechanism is also discussed. The core formula for the attention mechanism in U-Net is represented as follows (Zhang et al., 2019):

$$f(x) = W_f x, g(x) = W_g x, where\ x \in R^{C \times N} \quad (4)$$

$$\beta_{j,i} = \frac{\exp(s_{ij})}{\sum_{i=1}^{N} \exp(s_{ij})}, \text{where } s_{ij} = f(x_i)^T g(x_j) \quad (5)$$

$$o_j = v\left(\sum_{i=1}^{N} \beta_{j,i} h(x_i)\right), h(x_i) = W_h x_i, v(x_i) = W_v x_i \quad (6)$$

where $W_f, W_g, W_h \in R^{C'\times C}$, $W_v \in R^{C \times C'}$ are learnable 1×1 convolutional weight matrices. According to previous research (Zhang et al., 2019), $C'$ is set to $C/8$. The term $\beta_{i,j}$ represents the relevance of the $i$-th region to the $j$-th region, $C$ denotes the number of channels, and $N$ indicates the number of features in the hidden layer of the previous stage.

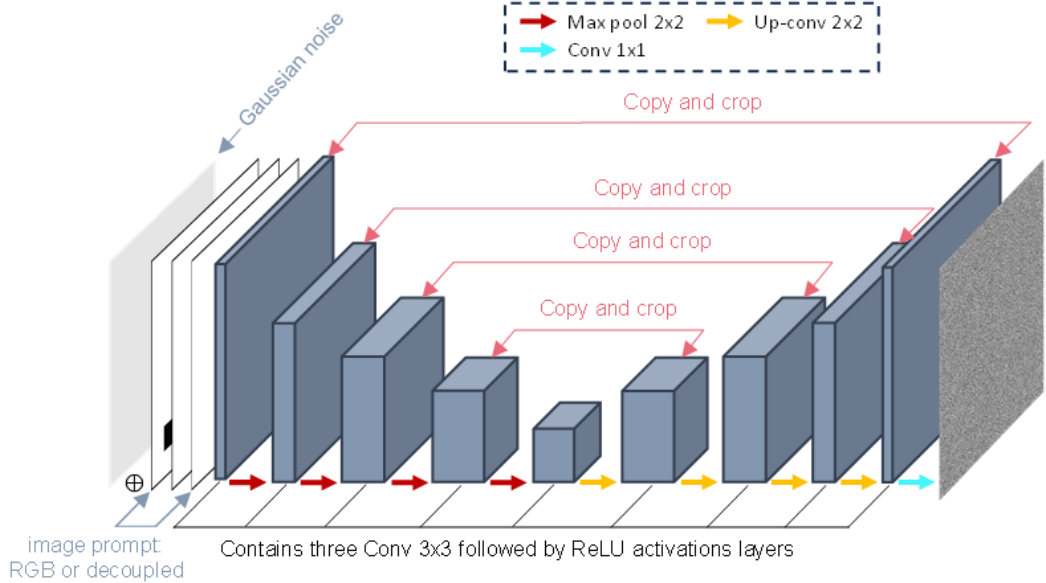

Figure 5 The architecture of U-Net

### 3.4 Validation and verification

This section analyzes and evaluates the proposed method against commonly used feature representations and other state-of-the-art model architectures to verify its superiority in evacuation simulation (Section 4). Finally, we demonstrate the practical application value of DiffEvac through case studies in Section 5.

#### 3.4.1 Comparison of different feature representations

We compared the proposed decoupled feature representation with the widely used RGB three-channel image representation to demonstrate its effectiveness.

The RGB three-channel image feature representation is the most straightforward approach. Currently, the RGB three-channel image format is widely used for building image generation models, where different building components (Liao et al., 2021; Lu et al., 2022; Fei et al., 2022) and architectural spaces (Nauata et al., 2020; Nourkojouri et al., 2023) are encoded and labeled using distinct RGB colors. If this input method is chosen, the room functional layout drawings processed in Section 3.1 can be directly used as input.

**3.4.2 Comparative analysis of different generative models**

Previous research has only used the pix2pix model (Isola et al., 2017), one of the most classic models in the GAN algorithm, to address this problem (Nourkojouri et al., 2023). Despite the remarkable image generation capabilities of various deep learning models, there has been no comprehensive analysis of their performance differences for this specific task. Therefore, we selected and compared several state-of-the-art models to identify the most suitable one for this application.

This study initially selected the conventional deep generative neural network model, U-Net (Ronneberger et al., 2015), as shown in Figure 5. U-Net features an encoder-decoder architecture with skip connections. The encoder part consists of convolutional layers and pooling layers for feature extraction, while the decoder part comprises convolutional layers and upsampling for generating segmentation masks. The skip connections (depicted as Copy and crop in Figure 5) facilitate the transfer of information between the encoder and decoder, helping to preserve spatial resolution and improve segmentation performance.

Additionally, this study also considers models from the GAN framework, which consists of a Generator and a Discriminator. The core concept of GANs is that the Generator creates images based on input data, while the Discriminator assesses whether these images are real or generated. Through adversarial training, the two networks are optimized to reach a balance (Goodfellow et al., 2014). The architecture of GAN is illustrated in Figure 6. In image generation algorithms, the Generator of a GAN network typically comprises convolutional and deconvolutional layers, similar to the U-Net structure shown in Figure 5. The Discriminator, on the other hand, generally consists of convolutional layers. Research by Liao et al. (Liao et al., 2021) indicates that the pix2pix algorithm (Isola et al., 2017) and the pix2pixHD algorithm (Wang et al., 2018) within the GAN framework outperform models like U-Net in shear wall generation tasks. This advantage is attributed to the "structural loss" in these algorithms, which aids in capturing the spatial distribution of shear wall layouts and reflecting the physical relationships between pixels, thus enhancing the effectiveness of shear wall generation. Given this similarity to the current task, this study will also compare the classic pix2pix and pix2pixHD algorithms within the GAN framework.

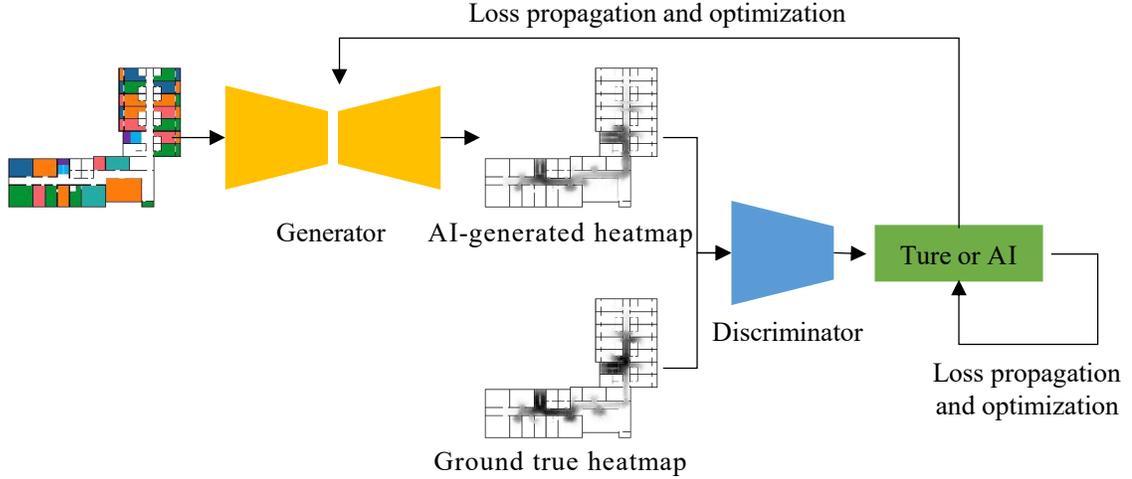

Figure 6 The architecture of GAN

### 3.4.3 Evaluation metrics

This study employs three commonly used image similarity assessment metrics to evaluate the similarity between the evacuation heatmaps generated by the model and the ground truth. These metrics include NRMSE (Normalized Root Mean Square Error), SSIM (Structural Similarity Index Measure), and PSNR (Peak Signal-to-Noise Ratio), which assess image similarity from different perspectives (Nourkojouri et al., 2023; Sara et al., 2019).

NRMSE measures the root mean square error between two images and normalizes it to a range between 0 and 1, as calculated in Equation 4. A value of NRMSE closer to 0 indicates greater similarity between the two images and better performance in the evacuation simulation. Additionally, SSIM takes into account three aspects: luminance, contrast, and structure, to simulate human perception of images, as calculated in Equation 5. The SSIM value typically ranges from -1 to 1, where 1 indicates that the two images are identical, 0 signifies no structural similarity between the images, and -1 indicates that the images are completely different. Finally, PSNR is an indicator used to assess image quality by comparing the peak signal-to-noise ratio between the original and distorted images, with higher values indicating better image quality, as calculated in Equation 6.

$$\text{NRMSE} = \frac{\sqrt{\frac{1}{N}\sum_{i=1}^{N}\left(I_{\text{model}}(i) - I_{\text{ground truth}}(i)\right)^2}}{I_{\max} - I_{\min}} \quad (7)$$

$$\text{SSIM} = \frac{\left(2\mu_{I_{\text{ground truth}}}\mu_{I_{\text{model}}} + C_1\right)\left(2\sigma_{I_{\text{ground truth}}I_{\text{model}}} + C_2\right)}{\left(\mu_{I_{\text{ground truth}}}^2 + \mu_{I_{\text{model}}}^2 + C_1\right)\left(\sigma_{I_{\text{ground truth}}}^2 + \sigma_{I_{\text{model}}}^2 + C_2\right)} \quad (8)$$

$$\text{PSNR} = 10 \cdot \log_{10}\left(\frac{I_{\max}^2}{\frac{1}{N}\sum_{i=1}^{N}\left(I_{\text{model}}(i) - I_{\text{ground truth}}(i)\right)^2}\right) \quad (9)$$

where $I_{\text{model}}(i)$ and $I_{\text{ground truth}}(i)$ represent the pixel values of the model-generated and ground truth images, respectively. $N$ is the total number of pixels, $I_{\max}$ and $I_{\min}$ are the maximum and minimum values in the images. $\mu_{I\max}$ and $\mu_{I\min}$ are the mean values of the two images, $\sigma^2_{I\text{model}}$ and $\sigma^2_{I\text{ground truth}}$ are their variances, $\sigma_{I\text{ground truth}/\text{model}}$ is the covariance. $C_1$ and $C_2$ are small constants added for stability.

### 3.4.4 Application in plan optimization

We illustrates the practical application value of the proposed study during the schematic design stages, where only architectural sketches are available, through an example of plan optimization via rapid evacuation simulation. The layout is optimized by only changing the position and number of evacuation doors while keeping the functional zones and room contours unchanged. It is important to note that the layout used for analysis was not included in the model training, meaning that the trained model had not previously encountered this specific layout.

To further illustrate the efficiency and accuracy of the proposed method, we also performed simulations using Pathfinder software and compared the prediction results and time required for both methods.

## 4 Experiments and Results

This section conducts a series of experiments to analyze the proposed DiffEvac against commonly used feature representations and other state-of-the-art model architectures to verify its superiority in evacuation simulation.

### 4.1 Experiment settings

The processed dataset is first divided into training, validation, and test sets in a ratio of 8:1:1. The training set is used to train and adjust the model parameters. The validation set is employed to select the optimal model, with the final model for testing being the one that performs best on the validation set. The test set is used for the final evaluation of the results. It is important to note that, to validate the model's generalization performance, all drawings in the test set are not present in the training and validation sets.

To evaluate the performance differences between the proposed method and alternative approaches, a series of 10 experiments was conducted. The specific experimental parameters are detailed in Table 2. Among them, the second column lists the name of each training experiment, which is composed of the model name, feature representations, and whether the attention mechanism is utilized. If the attention mechanism is employed, the suffix "-Att" is added to the experiment name. For instance, an experiment using decoupled feature representation with a diffusion model is labeled as "D-F-Att." The third column specifies the model employed, while the fourth column indicates whether RGB three-channel images (RGB) or decoupled representation (Feature) were used. The final column shows whether the attention mechanism was applied.

Table 2 Experimental IDs and their corresponding parameter configurations

| Group | ID | Model | Feature Representation | Attention |
|---|---|---|---|---|
| 1 | U-R | Unet | RGB | No |
| 2 | U-F | Unet | Feature | No |
| 3 | P-R | pix2pix | RGB | No |
| 4 | P-F | pix2pix | Feature | No |
| 5 | PH-R | pix2pixHD | RGB | No |
| 6 | PH-F | pix2pixHD | Feature | No |
| 7 | D-R | diffusion model | RGB | No |
| 8 | D-F (DiffEvac) | diffusion model | Feature | No |
| 9 | D-R-Att | diffusion model | RGB | Yes |
| 10 | D-F-Att | diffusion model | Feature | Yes |

For the pix2pix, pix2pixHD, and U-Net models, hyperparameters were selected based on prior research. Validation was performed every 50 epochs, and the best model was saved (Liao et al., 2021). The impact of different learning rates (0.0002, 0.0003, 0.0005, and 0.001) was also considered. For the diffusion model, hyperparameters were set following the guidelines provided by Gu et al. (Gu et al., 2024), with validation carried out every 100 epochs and the best model being saved.

The computing platform used for the experiments is configured as follows: Windows Server 2019 Standard as the operating system, an Intel Xeon E5-2682 v4 CPU running at 64 cores with a clock speed of 2.5 GHz, 54 GB of RAM, and an NVIDIA GeForce RTX 3090 GPU with 24 GB of memory.

## 4.2 Experiment results

The results of the experiments are shown in Table 3, which also includes the test results from existing research as benchmarks (Nourkojouri et al., 2023). Notably, it can be found that all results of Groups 1-10 in this study surpass those of previous research, demonstrating our models' strong performance. Additionally, the findings from the various experiments can yield the following conclusions:

Firstly, group "D-F" results show that DiffEvac, the proposed diffusion model combined with decoupled feature representation performs exceptionally well across all three metrics, achieving the overall best performance. At most, it outperforms previously proposed methods by up to 37.6% on SSIM and 142% on PSNR (Nourkojouri et al., 2023).

Secondly, based on the overall results, the performance of the different architectures is ranked as follows: diffusion > U-Net > pix2pix > pix2pixHD. Notably, U-Net outperforms the two GAN-based models. This may be attributed to the fact that one of the objectives of the discriminator in GAN architectures is to avoid mode collapse, thereby generating more diverse images. However, in this study, with a fixed room layout input, there is a unique and determined solution. Thus, the presence of a discriminator may not be as effective in producing the determined solution as the U-Net model.

Thirdly, comparing "D-R" with "D-R-Att" and "D-F" with "D-F-Att," it is evident that the impact of the attention mechanism is not substantial. This may be attributed to the high level of abstraction introduced by the attention layers in the model. Specifically, the network size at the input layer is 256, while the size at the attention layer is reduced to 256/8 = 32. At this level of abstraction, the model may struggle to capture fine details of the image, such as walls and exits, thereby diminishing the effectiveness of the attention mechanism. Additionally, incorporating attention mechanisms at lower resolutions will exceed the 24GB memory limit of current consumer-grade GPUs (e.g., RTX 3090, RTX 4090), making them less feasible for such hardware. Additionally, the generation of evacuation heatmaps relies more on global context rather than local feature correlations, which may further explain the limited impact of the attention mechanism.

Fourthly, the comparison between "D-R" and "D-F," as well as "D-R-Att" and "D-F-Att," reveals that for the diffusion model, using decoupled feature representation yields better results than using direct RGB inputs. However, this effect is not observed in the U-Net, pix2pix, or pix2pixHD architectures.

Finally, the superior performance of "D-F" may be also attributed to using single-channel grayscale evacuation heatmaps as the model output compared to using RGB evacuation heatmaps. This is because single-channel outputs reduce the number of model parameters, which is conducive to model learning, and thus achieves better results.

Table 3 The results of the experiments

| Group | ID  | Mean NRMSE | Mean SSIM | Mean PSNR |
|-------|-----|------------|-----------|-----------|
| 1     | U-R | 0.1074     | 0.9120    | 20.35     |
| 2     | U-F | 0.1089     | 0.9096    | 20.21     |
| 3 | P-R | 0.1138 | 0.9002 | 19.79 |
| 4 | P-F | 0.1155 | 0.9030 | 19.62 |
| 5 | PH-R | 0.1175 | 0.8980 | 19.38 |
| 6 | PH-F | 0.1194 | 0.8943 | 19.22 |
| 7 | D-R | 0.0709 | 0.9618 | 23.88 |
| 8 | D-F (DiffEvac) | **0.0681** | **0.9632** | **24.20** |
| 9 | D-R-Att | 0.0704 | 0.9600 | 23.99 |
| 10 | D-F-Att | 0.0694 | 0.9613 | 24.13 |
| 11 | Test 1 (Nourkojouri et al., 2023) | / | 0.8900 | 18.74 |
| 12 | Test 2 (Nourkojouri et al., 2023) | / | 0.7000 | 10.00 |

Where Test 1 refers to the entire test set, while Test 2 denotes a subset of Test 1, specifically comprising rooms from the test set that appear less frequently in the training set.

Evacuation heatmaps generated by different models for a typical case are shown in Figure 7. To highlight their differences from the ground truth, pixels with significant grayscale differences are displayed in distinct colors. Notably, blue areas represent locations where crowd density is underestimated by half compared to the ground truth, while red areas indicate overestimation by the same margin. Additionally, the areas with the greatest discrepancies have been magnified to emphasize the performance differences between the models. The results from models with attention mechanisms are not displayed due to their minimal impact. From the generated results, it is evident that Unet, pix2pix, and pix2pixHD models exhibit issues such as inaccurately overestimating crowd density. In contrast, DiffEvac generates results that are closest to the actual evacuation heatmaps, demonstrating the best overall performance. The similarity of the generated heatmaps for this typical case was assessed using computer vision metrics NRMSE, SSIM, and PSNR, as shown in Table 4. It is evident that the images generated by DiffEvac achieve the highest performance across all three metrics.

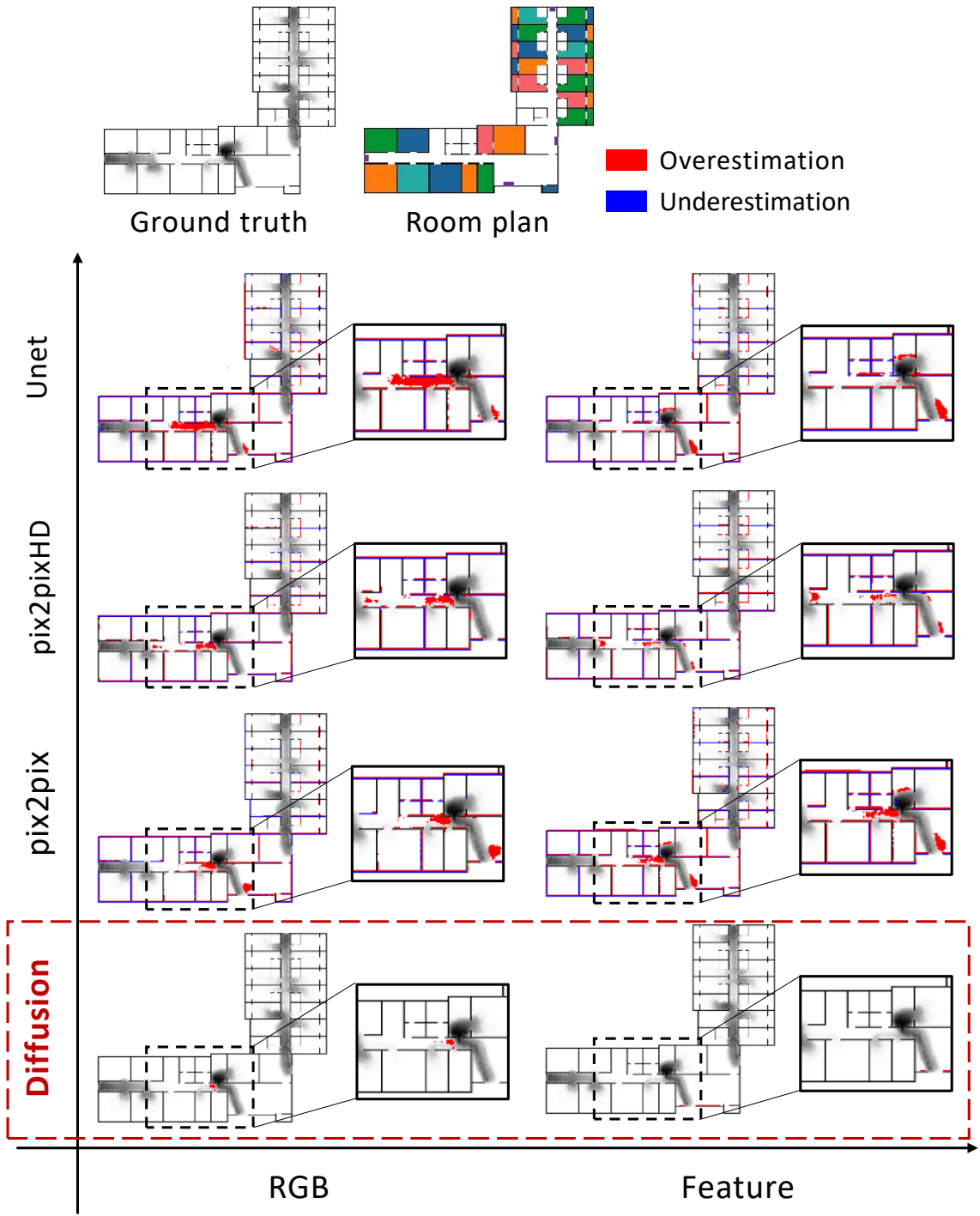

Figure 7 Typical evacuation heatmaps generated by different models

Where "U-" denotes the U-Net model, "P-" denotes the pix2pix model, "PH-" denotes the pix2pixHD model, "D-" denotes the diffusion model, "F" indicates the use of the decoupled representation as input, "R" indicates the use of RGB three-channel images as input and blue areas represent locations where crowd density is underestimated by half compared to the ground truth, while red areas indicate overestimation by the same margin.

Table 4 The results of the typical case shown in Figure 7

| Group | ID | NRMSE | SSIM | PSNR |
|---|---|---|---|---|
| 1 | P-F | 0.8737 | 0.5721 | 10.09 |
| 2 | P-R | 0.8966 | 0.5580 | 9.872 |
| 3 | PH-F | 0.7643 | 0.6169 | 11.25 |
| 4 | PH-R | 0.7678 | 0.6198 | 11.22 |
| 5 | U-F | 0.8552 | 0.5867 | 10.28 |
| 6 | U-R | 0.8930 | 0.5591 | 9.907 |
| 7 | D-R | 0.6151 | 0.7454 | 13.14 |
| 8 | D-F (DiffEvac) | **0.5557** | **0.7733** | **14.02** |

Note that a lower NRMSE value indicates better performance and greater similarity between the two images. An SSIM value closer to 1 reflects higher image similarity, while a higher PSNR value indicates better image quality.

In addition, this study compared the performance differences of DiffEvac across different layout types and contrasted it with the pix2pix model of the GAN series used in Nourkojouri et al. (2023). The results are shown in Figure 8, with the complexity of building layouts increasing from top to bottom. The results indicate that as layout complexity increases, the performance of the pix2pix model in evacuation simulations deteriorates, while the method proposed in this study demonstrates strong adaptability and stability across different levels of complexity. This highlights the superior generalization ability of the proposed method, proving its advantage and reliability in evacuation simulation tasks.

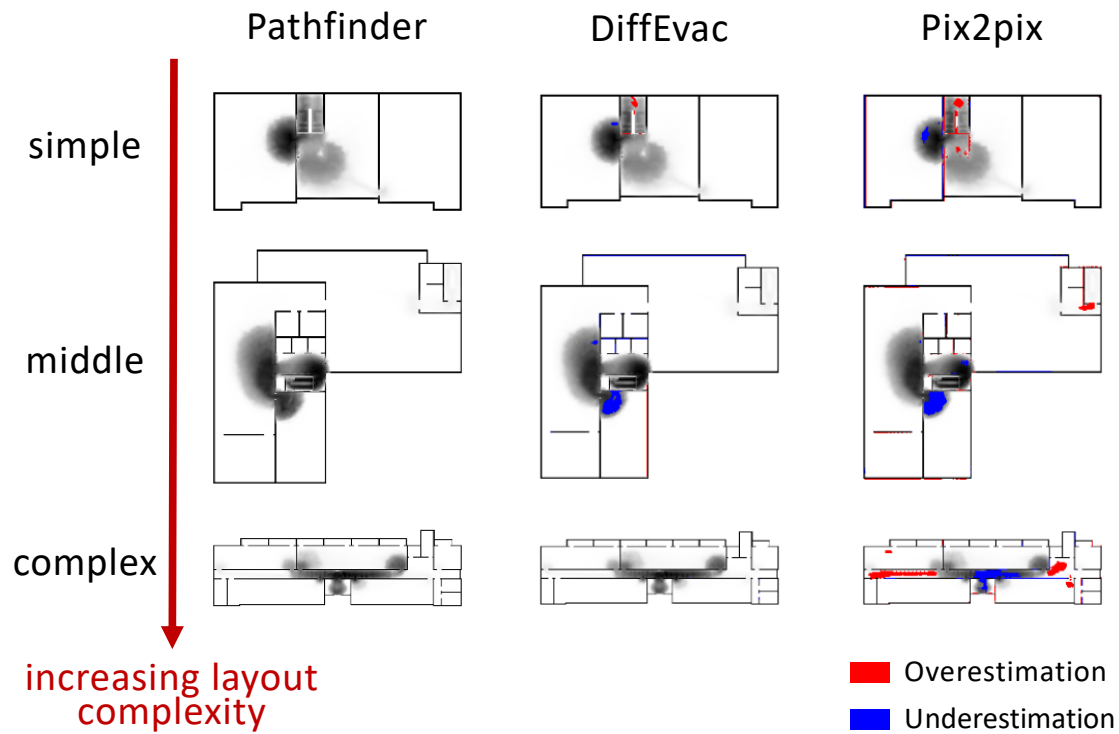

Figure 8 Results for layouts with different levels of complexity

## 5 Case Study

The room layout before optimization is shown in Figure 9(a). This layout was input into DiffEvac, described as the "D-F" model in Section 4, and the predicted evacuation heatmap for the room was generated within 74 seconds, as shown in Figure 9(c). It can be observed that the crowd evacuates through the exit at the bottom of the room, causing significant congestion as many people must traverse the corridor leading to this exit. Following optimization, an additional exit was introduced on the right side of the room, as shown in Figure 9(d). Using the same method, the evacuation heatmap was re-generated, as depicted in Figure 9(f). The updated layout demonstrates that with the new exit, the crowd on the right side now uses this additional exit, effectively reducing the evacuation pressure on the central corridor.

Additionally, Pathfinder was used to model the room layout and generate evacuation heatmaps for comparison, as shown in Figures 9(b) and 9(e). Comparing Figures 9(b) with 9(c) and Figures 9(e) with 9(f), it is evident that the proposed surrogate model can predict the heatmaps accurately. The process of modeling and simulation using Pathfinder takes approximately 20 minutes (1200 seconds), which is not suitable for the rapid iteration and optimization required during the schematic design stage.

In contrast, the proposed evacuation surrogate model generates the evacuation heatmap in just 74 seconds, achieving nearly a 16-fold improvement in analysis efficiency. It is important to note that as the complexity of the room increases, the modeling time with Pathfinder also increases significantly. However, the time required for the proposed surrogate model remains nearly constant, thus significantly improving simulation efficiency and meeting the need for evacuation adjustments during the design phase.

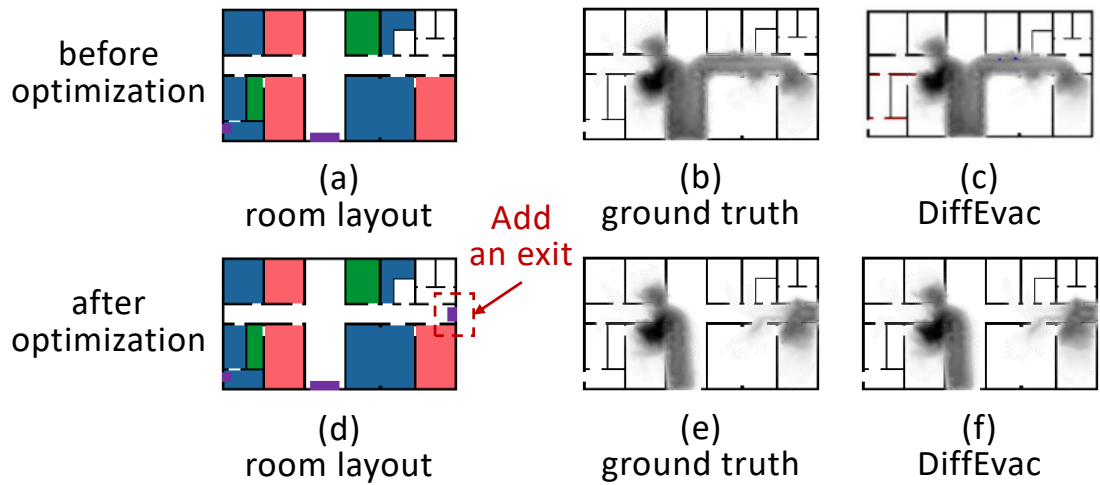

Figure 9 Plan optimization via rapid evacuation evaluation

In addition, this study compared the evacuation simulation times of DiffEvac and Pathfinder for the layouts with varying complexity depicted in Figure 8, with the results presented in Figure 10. As layout complexity increases, Pathfinder's simulation time rises accordingly. In contrast, the proposed method, DiffEvac, maintains robust stability in prediction time across different complexity levels and consistently outperforms Pathfinder, being approximately 20 times faster. This demonstrates the superior efficiency and adaptability of the proposed method, making it a more suitable choice for real-time evacuation simulations in complex architectural scenarios.

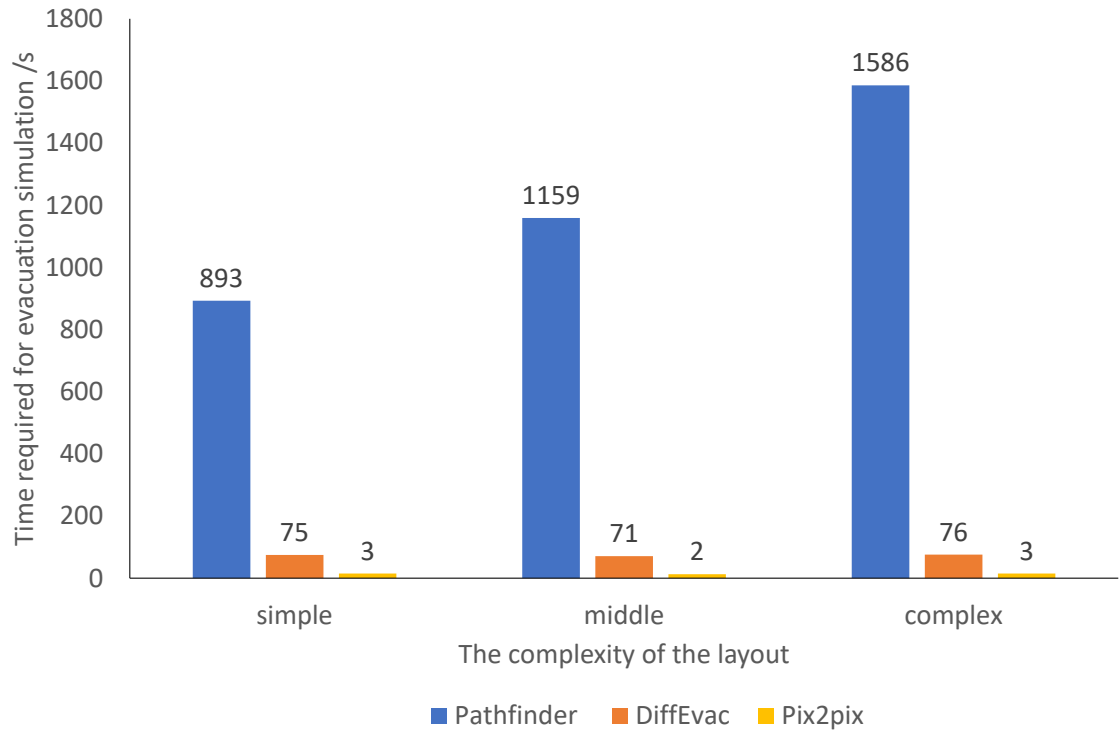

Figure 10 Time required for layouts with different levels of complexity

## 6 Conclusion

During the review and adjustment process, architects often face numerous design alternatives. Efficiently considering the impact of building layout on evacuation in the early design stage can dramatically reduce the frequency of revisions and rework caused by improper evacuation plans. Therefore, this study proposes an efficient evacuation simulation method based on Generative Models, designed to learn and simulate building evacuation patterns. This method enables designers to rapidly iterate and adjust building layouts during both the initial design phase and post-review modification stage, thus enhancing safety design. Specifically, 81 office building floor plans were collected, and, through annotation based on relevant regulations and data augmentation, 399 room functional layout drawings were generated to develop the algorithm. Pathfinder was then used to model these layouts and produce 399 corresponding evacuation heatmaps. Based on these, a diffusion model was proposed to establish a rapid evacuation evaluation model. Finally, to improve the accuracy of the rapid evacuation simulation method, this study systematically compares different deep learning models and feature representations. The main conclusions are as follows:

(1) Overall, the proposed DiffEvac demonstrated the best performance. Compared to existing research, DiffEvac achieves up to a 37.6% improvement in SSIM and a 142% improvement in PSNR.

Therefore, it is recommended that the diffusion model combined with decoupled feature representation be used for further development in evacuation surrogate model research.

(2) This method allows for the rapid generation of evacuation heatmaps from room functional layout drawings within 2 minutes, achieving a 16-fold increase in efficiency compared to traditional simulation approaches. These significant improvements enable quick iterations and adjustments during both the design and review stages.

By learning evacuation heatmaps generated by Pathfinder software in the dataset, the model grasps the patterns of refined evacuation simulations, enabling it to replace Pathfinder and quickly generate simulation results. The proposed method provides a promising alternative to traditional simulation tools, reducing computational time while maintaining the accuracy and reliability of evacuation predictions. However, this study still has limitations that could be addressed in future research. Specifically, the evacuation surrogate model is currently validated only for office buildings, and future work could extend it to other building types. Additionally, simplifications were made by excluding factors such as personnel speed, response time, disaster dynamics, and exit variations, which should be considered in future studies to enhance the model's applicability.

## Acknowledgement

The authors are grateful for the financial support received from the National Natural Science Foundation of China (Nos. 52238011, 52378306).